\documentclass[conference]{IEEEtran}
\IEEEoverridecommandlockouts
\usepackage{multirow} 
\usepackage{booktabs} 
\usepackage{caption}
\usepackage{subcaption}

\usepackage{cite}
\usepackage{amsmath,amssymb,amsfonts}
\usepackage{algorithmic}
\usepackage{graphicx}
\usepackage{textcomp}
\usepackage{xcolor}
\def\BibTeX{{\rm B\kern-.05em{\sc i\kern-.025em b}\kern-.08em
    T\kern-.1667em\lower.7ex\hbox{E}\kern-.125emX}}
\begin{document}

\title{Attention, Anomalies! Handling Attention Layers in Unsupervised Federated Outlier Detection}
\author{
\IEEEauthorblockN{
Mihailo Ilić\IEEEauthorrefmark{1},
Miloš Savić\IEEEauthorrefmark{1},
Vladimir Kurbalija\IEEEauthorrefmark{1},
Mirjana Ivanović\IEEEauthorrefmark{1},
Giancarlo Fortino\IEEEauthorrefmark{2},
Dušan Jakovetić\IEEEauthorrefmark{1}
}

\\

\IEEEauthorblockA{\IEEEauthorrefmark{1}%
Department of Mathematics and Informatics, Faculty of Sciences, University of Novi Sad, Serbia}
\{milic,svc,kurba,mira,dusan.jakovetic\}@dmi.uns.ac.rs
\IEEEauthorblockA{\IEEEauthorrefmark{2}%
Department of Informatics, Modeling, Electronics, and Systems, University of Calabria, Italy}
giancarlo.fortino@unical.it
}

\maketitle

\begin{abstract}
Attention layers are the backbone of today's most powerful and impactful models. Models with multi-million and billion parameters rely on contextual knowledge provided by attention layers. However, their use goes well beyond just being the core component of large language models. One particularly interesting application is in Memory Augmented Autoencoders (MemAE), specifically for unsupervised representation learning in outlier detection tasks. It was shown that attention helps these models be more effective in centralized learning scenarios. Our work aims to address the lack of specialized aggregation techniques in Federated Learning (FL) when it comes to MemAE models. In this paper we analyze the intricacies of the architecture behind Memory Augmented Autoencoders, and propose novel, guided approaches to effectively aggregate these models in federated scenarios. We demonstrate our approach on non-IID datasets and show that these novel aggregation schemes are more robust when dealing with numerous edge nodes in environments with unbalanced datasets, specifically for unsupervised anomaly detection scenarios. This approach improves the performance of even very shallow autoencoders, allowing them to be used in resource constrained environments. 

\end{abstract}


\section{Introduction}

Federated Learning (FL) has quickly grown into one of the most widely-used and researched learning paradigms. This is a result of intensified distributed data collection~\cite{10.1561/2200000083} in various scenarios, most commonly in Internet of Things (IoT)~\cite{pmlr-v54-mcmahan17a} and collaborative learning on sensitive data, like use cases commonly found in medicine~\cite{ilic2024towards,10.1007/978-3-030-11723-8_9}. 

Various benefits come from collecting data and optimizing models at the edge, and coordinating the learning process by a central server. Some of them include: eliminated dataset communication overhead, eliminated centralized big data storage concerns, increased data privacy and security, reduced computational costs (from distributed computation), etc.

A key topic in FL are the aggregation functions used by the server which dictate how individual model updates are to be incorporated into the next iteration of the global model. Some general purpose aggregation functions are FedAvg~\cite{pmlr-v54-mcmahan17a}, FedProx~\cite{MLSYS2020_1f5fe839}, and FedHybrid~\cite{niu2023fedhybrid}. The choice of aggregation function  greatly depends on the use case and the intrinsics of the model being trained. Specialized models and scenarios require aggregation functions tailored to their unique requirements, which usually are modifications of the general purpose functions. 

Massively distributed learning environments like those in IoT suffer from statistical differences by data collected by each individual edge node. Data drifts present across the pool of edge nodes are a challenge in training shared global models.
A balance needs to be struck between good generalizability of the model, i.e. good performance on average across the clients, and maximising performance on each individual edge node. These types of datasets are non-independent and identically distributed (non-IID), and present a challenge when training distributed models. Each client (edge node) can be viewed as its own learning task, and the challenge can then be viewed as training on multiple different learning tasks.

Representation learning (RL) is one of the proposed methods to navigate the issue of multi-task learning. The goal is training machine learning (ML) models able to extract relevant features from data and focus on those to solve the underlying ML task (e.g. classification or regression). One popular approach in RL is the use of autoencoder networks, neural networks with two distinct parts: an encoder and decoder. The encoder learns a latent representation (encoding) of input data points, i.e. they project the input points to a vector of smaller dimensionality. Quality encodings are those that capture the important attributes and relationships in the input data and allow the decoder to reconstruct the original data from them. 
 
Memory-augmented autoencoders (MemAE)~\cite{Gong_2019_ICCV} are a special kind of autoencoder that utilize an attention mechanism between the encoder and decoder, specifically designed for anomaly detection (AD) tasks. Training is done in an unsupervised fashion, where the model is only shown normal data points.
The attention mechanism allows them to remember prototypical examples from the training data and enables them to reconstruct normal data samples well, but struggle with out-of-sample (outlier) data points and produce a significant error when reconstructing them. 

Federated anomaly detection relates to both detecting anomalous clients~\cite{DBLP:journals/access/XiaCYM23,DBLP:journals/air/ZhangYMN24}, and anomalous data points found within client datasets. The first case focuses on the detection of malicious clients, or those that negatively impact the performance of the global model. This paper focuses on the second type, where the goal is detecting anomalous data points collected by federated clients.

The MemAE model has been shown to produce very good results in centralized AD scenarios~\cite{Gong_2019_ICCV}, and has been tested in federated settings as well~\cite{10839838,11336743}. Even though it showed good results compared to other models for federated anomaly detection, it has been pointed out in~\cite{10839838} that no specialized aggregation function exists for MemAE which accounts for its architectural differences, namely the attention layer. A stochastic aggregation function based on FedProx was proposed in~\cite{11336743}, that takes into account these structural intricacies of the MemAE model. This approach showed better performance compared to baseline methods, but requires further validation on a broader collection of datasets.

The main contributions of this paper are:

\begin{enumerate}
    \item We validate the stochastic MemAE aggregation function on a broader set of both IID and non-IID datasets.
    \item We introduce three novel, guided approaches to aggregating the MemAE model based on K-Means clustering, K-Medoids clustering, and Facility Location, taking into account the specifics of the memory layer.
    \item The newly-proposed methods show an improvement over existing ones based on standard and stochastic FedAvg aggregation of MemAE.
    \item These methods show good results on shallow models, making them a good fit for IoT scenarios.
\end{enumerate}

The rest of the paper is structured as follows. Section~\ref{sec:rw} gives an overview of related work. Architectural details of the MemAE model and novel aggregation strategies are presented in Section~\ref{sec:methodology}. The experimental work is presented in Section~\ref{sec:experiments} and results are discussed in Section~\ref{sec:results}, followed by concluding remarks in the last section.

\section{Related Work}
\label{sec:rw}

Anomaly detection is the process of determining whether a data point or a subset of data points is normal or anomalous. They fall into three different categories: point anomalies, contextual anomalies, and collective anomalies~\cite{DBLP:journals/access/NassifAND21}. Point anomalies are unique and thus not in line with the rest of the data, contextual anomalies are only considered outliers in certain situations (contexts), while collective anomalies are a set of related data points which in a given context fall into the category of outliers. 

Machine learning (ML) approaches to anomaly detection (AD) are either supervised or unsupervised. Supervised AD is typically reduced to a classification problem that can be addressed by plenty methods, ranging from classic ML techniques (e.g, support vector machines, random forest, probabilistic classification) to modern deep learning solutions~\cite{10.1145/3439950}. However, supervised AD requires high quality and high volume training datasets with labeled anomalies, that are usually not available, especially in domains experiencing rapid technological developments such as IoT~\cite{9402912} and tax administration~\cite{SAVIC2022116409}. Since anomalies are rare events, the main challenge of supervised AD methods is the problem of class imbalance that can be tackled either at the data level (e.g., by undersampling or oversampling) or the algorithm level (e.g., by cost-sensitive ML algorithms). 

Considering above mentioned difficulties, more research attention and practical relevance is given to less-precise, but less-restrictive unsupervised AD approaches~\cite{10.1145/3381028}. The article by Ruff et al.~\cite{RufPIEEE21} emphasizes that there are two big classes of unsupervised AD algorithms, shallow and deep, that can be viewed from a unified perspective, belonging to one of the following three broad categories: (1) density estimation and probabilistic models, (2) one-class classification models and (3) reconstruction models. The focus of this paper is on the third category, reconstruction models, that are based on low-dimensional representation learning for normal data points, with autoencoders as the most dominant approach.

Detecting outliers in distributed settings like FL can relate to either detecting anomalous clients or anomalous data points within the datasets of particular clients. The first case is also viewed mostly through the lens of security, covering aspects of attacks and defense strategies in FL~\cite{DBLP:journals/tnn/LyuYMCSZYY24,DBLP:journals/access/NetoHDMF23}, where the goal is detecting which clients are ``poisoning'' the global model, meaning that they are submitting suboptimal weight updates either intentionally or on accident. The other scenario, detecting outliers in clients' datasets, is a different problem which is plagued by the non-IID nature of local datasets and specific data distributions~\cite{DBLP:journals/access/JithishAMY23,11336743,10839838,11336407}, through either supervised~\cite{DBLP:journals/access/JithishAMY23}, unsupervised~\cite{11336743}, or self-supervised methods~\cite{10566052}. This paper focuses on the second case, where the goal is training outlier detection models through FL, specifically using unsupervised methods and relying on representation learning models. 

Representation learning (RL) is the process of learning, often more compact, representations of the input data in a way which extracts relevant information suitable for training models for other purposes such as e.g. classification or regression~\cite{6472238}. Quality representations can help increase the performance of models for other downstream tasks. The motivation behind applying representation learning in federated settings~\cite{10.1145/3378679.3394530} comes from the problem of different data distributions across client datasets. Learning from different data distributions can be viewed as learning models for different tasks, and the fusion of these models is similar to multi-task learning~\cite{9492755}. Learning representations applicable to numerous clients with many different data distributions is called Federated Representation Learning (FRL)~\cite{10.1145/3378679.3394530}. 
 
Neural networks are models often used in FL settings since it is relatively straightforward to share and aggregate their weights in different ways. Autoencoders are a type of neural network architecture used for RL. They learn to represent input data points in a lower dimension, from which it is possible to accurately reconstruct the original input. These compact representations are then used for downstream tasks. Additionally, when trained in an unsupervised fashion, and being shown only inlier data points, autoencoders can be used for anomaly detection needs. The reconstruction loss in this case should be larger for previously unseen data points (anomalies), and by setting some threshold on the reconstruction error, points can be labeled inliers or outliers. The authors of~\cite{Gong_2019_ICCV} propose a special type of autoencoder architecture, called MemAE, that takes advantage of attention mechanisms by introducing a memory layer in between the encoder and decoder parts of the network. This architecture shows clear advantages in centralized settings, and has been introduced to FL in~\cite{10839838,11336743}. However, these aggregation strategies either do not fully consider the intricacies of the memory layer, treating it as any other weight in the neural network~\cite{10839838}, or provide a stochastic way of aggregating the memory matrices of different edge nodes~\cite{11336743}. In this paper, we tackle these issues by introducing guided methods of aggregating the memory module based on clustering and coreset selection methods. 

Optimal client selection has been explored extensively for use cases in federated learning. Specifically, clustering methods based on submitted client updates have been used to group clients with similar updates to improve model personalization~\cite{9743558}. Some examples include the use of hierarchical clustering~\cite{DBLP:journals/www/LongXSZWJ23}, K-Means clustering~\cite{DBLP:journals/bigdatama/XiongZSWWLG24, 9644782}, convex clustering~\cite{DBLP:conf/isc2/ArmackiBJK22}.

Selecting a coreset of informative clients is another method of optimal client selection in federated learning. One such approach is through submodular optimization~\cite{10.1145/3638052},
a process of selecting a subset of the most diverse set of elements such that they retain the statistical distribution of the original dataset~\cite{krause2014submodular,NIPS2013_a1d50185}. A notable approach is the facility location~\cite{krause2014submodular} function, which has been used in federated learning for optimal client selection~\cite{10.1145/3638052,DBLP:conf/iclr/Balakrishnan0ZH22}. 

The methods we propose in this paper also rely on clustering and coreset selection methods like these. However, the key difference is that in our approach we do not consider the entire set of weights, nor do we apply these techniques specifically for client selection. Functions like K-Means and Facility Location are used to determine the optimal selection of vectors stored in the memory modules of the MemAE models submitted to the FL server by clients during training. 

\section{Methodology}
\label{sec:methodology}

The work presented in this paper is in the context of anomaly detection, using unsupervised representation learning methods on MemAE models in federated learning scenarios. 

Unsupervised representation learning can be accomplished in numerous ways, in this work we focus on autoencoder models. They map an input data point $x$ to a latent representation $z$, which is then decoded into $x'$ in an attempt to reconstruct the input $x$. The quality of the reconstruction can be measured in terms of the error (deviation) from the original input, called the reconstruction loss. As in~\cite{Gong_2019_ICCV,10839838,11336743}, in this paper we also rely on the squared $L2$ norm, as defined in Equation~\ref{eq:reconstruction_loss}.

\begin{equation}
    \label{eq:reconstruction_loss}
    e = ||x - x'||_2^2
\end{equation}

Unsupervised anomaly detection relies on the reconstruction loss. Anomaly detection models are trained so that they only see normal (inlier) data points, causing the model to have higher reconstruction errors $e$ for outliers. By setting a threshold $\hat{e}$ for the reconstruction loss, data points can be classified into inliers ($e < \hat{e}$) and outliers ($e \geq \hat{e}$).

In this paper we focus on training anomaly detection models in federated environments, specifically the MemAE model, by training in an unsupervised fashion on inlier data points.

\begin{figure}[h]
    \centering
    \includegraphics[width=0.5\textwidth]{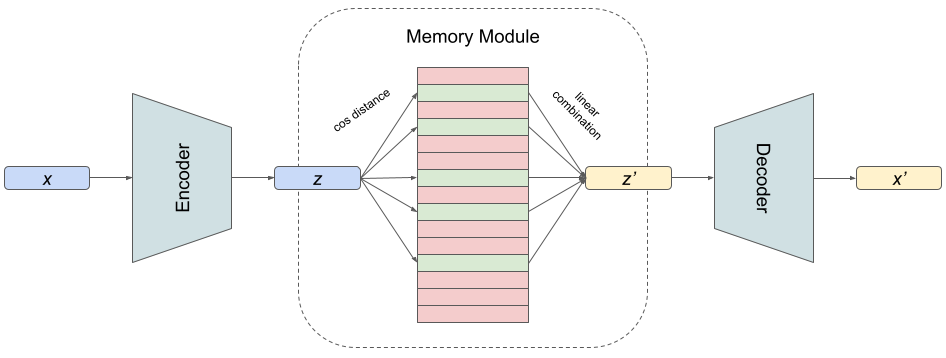}
    \caption{A diagram of the MemAE model. The encoder projects $x$ to a latent representation $z$. A new representation $z'$ is obtained from a linear combination of entries in the memory module, after which it is decoded into $x'$.}
    \label{fig:memae_arch}
\end{figure}

The MemAE model~\cite{Gong_2019_ICCV} is depicted in Figure~\ref{fig:memae_arch}. It incorporates attention by adding the memory matrix of dimension $N \times C$ in between the encoder and decoder. $N$ is the number of memory slots (rows) and $C$ is the dimension of the latent representations. The memory module learns $N$ prototypical examples from the training dataset. The input $x$ is mapped to an encoding $z$, after which the rows in memory are ranked by cosine similarity to $z$. A hard shrinkage operation is applied~\cite{Gong_2019_ICCV} to the similarity scores which induces sparsity in the memory. A new encoding $z'$ is obtained by a linear combination of the chosen prototypes from memory, which is then decoded into $x'$. In our experiments, the hard shrinkage threshold is set to $1 / N$, as proposed in~\cite{Gong_2019_ICCV}.

The use of MemAE in federated outlier detection showed promising results in~\cite{10839838}. However, in this work the memory module was treated equally to the rest of the network, and was aggregated by the general purpose FedProx~\cite{MLSYS2020_1f5fe839} function. The authors pointed out a need for specialized aggregation functions which take into account the architecture of the attention layer.

Since the attention layer stores prototypical vectors in its rows, a simple element-wise weighted average of the matrices does not take into account the internal structure of the attention layer of which the building blocks are rows and not individual cells. Because of this, a stochastic approach was proposed in~\cite{11336743}, involving a random row selection to form the new memory matrix. This approach outperformed the baseline FedProx aggregation, however, a need for a guided (non-stochastic) approach is still needed.

In this paper we propose the use of three different informed aggregations based on FedAvg. Regular network weights are aggregated without any deviation from FedAvg, while the memory module is aggregated using one of the following three approaches.

\textbf{Facility Location --} The attention aggregation is conducted via a general purpose facility location function based on cosine similarity scores~\cite{krause2014submodular} of the combined rows from all client matrices. These functions choose a set of representatives from the target dataset, such that they maximize the pairwise similarities between the representatives and the leftover points from the target dataset. The idea behind this is to chose $N$ representative examples which cover the original space well, from the $E \cdot N$ vectors received from $E$ clients.

\textbf{K-Means --} In this approach, $E$ clients each submit $N$ memory vectors, and the global memory matrix is formed from $N$ centroids provided by K-Means clustering. All $E \cdot N$ vectors are normalized prior to clustering, and the distance metric is also cosine similarity, to stay consistent with the similarity scoring applied in the attention mechanism of MemAE.

\textbf{K-Medoids --} This approach is similar to K-Means, but the new memory matrix consists of specific vectors submitted by clients rather than newly computed centroids.

\section{Experiments}
\label{sec:experiments}

\subsection{Datasets}

In order to verify the aggregation strategies presented in this paper, three different datasets were prepared for anomaly detection scenarios in federated settings. The datasets were constructed similarly to related work in~\cite{11336743} and~\cite{10839838}, with additional steps taken to verify the results in both IID and extremely non-IID scenarios, like those available in the LEAF benchmark datasets~\cite{caldas2019leafbenchmarkfederatedsettings}. All of the datasets presented in this paper were adapted to fit anomaly detection scenarios in federated settings. 

The first is the KDDCUP10 network intrusion dataset~\cite{kdd_cup_1999_data_130}, a 10\% subset of the original KDDCUP dataset. Having $494021$ data points, with $42$ different attributes (continual and categorical), it is a good candidate to test on a large number of edge nodes since it allows for extremely unbalanced (non-IID) scenarios. It contains data on normal network traffic and network intrusion attacks like denial of service, probing, and surveillance. Categorical columns were one-hot encoded and the entire dataset was min-max normalized. The four different attack classes were all merged into one anomaly class.

The NSL KDD dataset~\cite{tavallaee2009detailed} is an updated version of the KDD dataset. The removal of redundant records is the main difference between this dataset and the previous KDD dataset. The evaluation metrics of this dataset are considered more realistic, since bias in the data is removed which can inflate accuracy scores in some scenarios.

The PAMAP2 physical activity monitoring dataset~\cite{pamap2_physical_activity_monitoring_231} consists of data collected by wearable devices from $9$ subjects performing $18$ different physical activities. These include walking, lying, standing, cycling, and many more. For anomalous activities, ``car driving'', ``playing soccer'', and ``rope jumping'' were chosen arbitrarily. Min-max normalization was applied to all attributes as part of the training process, and the class labels were converted to $0$ (inliers) and $1$ (outliers).

Since the ML goal is unsupervised anomaly detection, the training datasets were constructed in such a way as to include only inlier data, having clients train models only on normal data points. The test dataset was constructed such that it consists of an equal number of inlier and outlier data points (a $1:1$ ratio), and was centralized for convenience, so that it can be used by the FL server after each training round. Inliers that went into the test set were chosen in a uniform random fashion from all of the inliers, and the rest of the normal data points were distributed between the federated clients. 

\subsection{Experimental Setup}
\label{sec:experimental-setup}

All of the experiments were conducted on $100$ edge nodes with full participation in each round.
A greater number of edge nodes produces a larger variety of feature vectors stored in the rows of each client's MemAE memory matrix.
Consequently, this makes the selection of the ``right'' prototypical examples from the attention layer more challenging. 

\begin{figure*}[t]
    \centering
    \begin{minipage}[t]{0.5\textwidth}
        \centering
        \includegraphics[width=\textwidth]{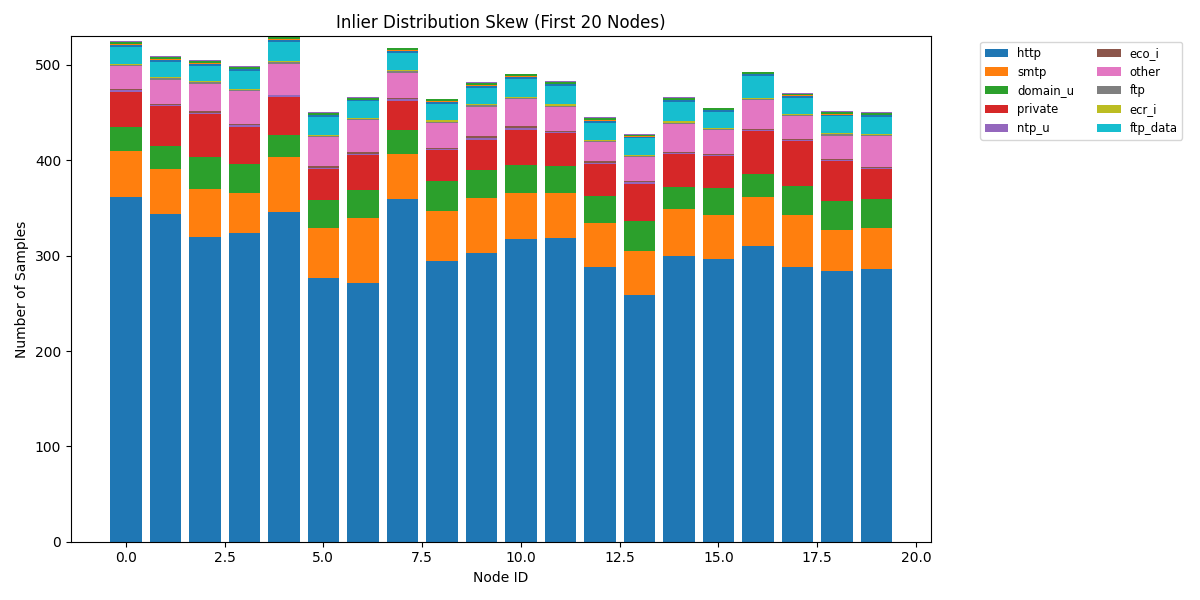}\\
        \small (a) IID setting ($\alpha = 100$)
    \end{minipage}\hfill
    \begin{minipage}[t]{0.5\textwidth}
        \centering
        \includegraphics[width=\textwidth]{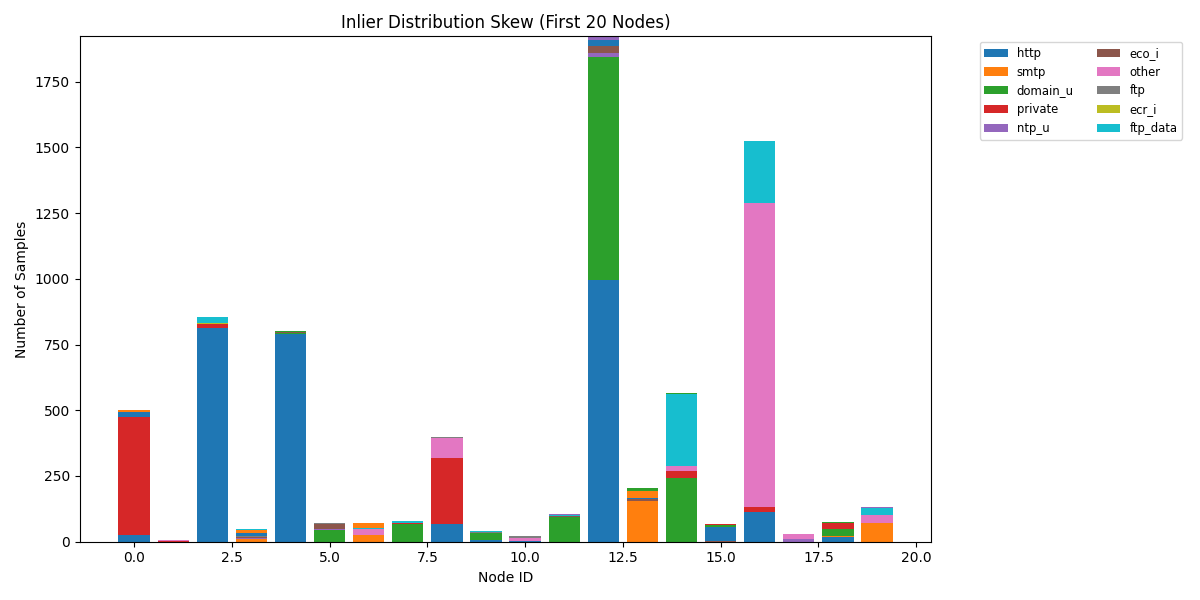}\\
        \small (b) Non-IID setting ($\alpha = 0.1$)
    \end{minipage}
    \caption{Examples of client data distributions of the KDD dataset for IID and non-IID scenarios.}
    \label{fig:dirichlet-client-distributions}
\end{figure*}

In order to verify the robustness of the proposed methods, the experiments included both IID and extremely non-IID scenarios. This was achieved by sampling data from a Dirichlet distribution~\cite{hsu2019measuringeffectsnonidenticaldata} over the dataset attributes. For the KDD and NSL KDD datasets, the communication protocol (HTTP, SMTP, etc.) was chosen as the partition column, while for the PAMAP2 dataset this was the activity type. By setting the $\alpha$ parameter, either balanced or extremely unbalanced data distributions across edge nodes can be obtained. In our experiments, we chose $\alpha = 100$ for the IID scenario and $\alpha = 0.1$ for the non-IID scenario. A visual example of the client data distribution can be seen in Figure~\ref{fig:dirichlet-client-distributions}.

The experiments included multiple models and aggregation strategies. To verify our newly-proposed methods, we compare them to two baselines: the first being a regular autoencoder, and the second being just a regular weight aggregation strategy applied to the entire MemAE model, like the one mentioned in~\cite{10839838}. The second baseline relies on FedAvg~\cite{pmlr-v54-mcmahan17a}, and aggregates the weight matrices exactly the same as done for regular network weights in the encoder and decoder. All the different models will be referred to as follows:

\textbf{Autoencoder --} A regular autoencoder model which serves as the first baseline model.

\textbf{MemAE --} The MemAE model aggregated via regular FedAvg. No special handling of the memory layer was applied here, and this approach serves as the second baseline.

\textbf{Random --} The MemAE model where the memory module was aggregated stochastically like in~\cite{11336743}.

\textbf{Facility Location --} The aggregation of the memory module was conducted after each global training round using a general purpose facility location function based on the cosine similarity scores, using the implementation in~\cite{JMLR:v21:19-467}. The rest of the weights are aggregated by regular FedAvg.

\textbf{K-Means --} Memory vectors are aggregated after each training round using K-Means clustering, while the rest of the weights are handled using FedAvg. Before applying K-Means, all memory vectors are normalized. The output memory vectors from this approach are new prototypical examples,
the cluster centers produced by K-Means.

\textbf{K-Medoids --} Similar to the previous approach, just with K-Medoids clustering. The output memory matrix from this approach contains exact vectors from the original dataset.

All experiments used a shallow autoencoder architecture, having one hidden layer in both the encoder and decoder, with minimal deviations from each other between datasets. The encoder's hidden layer had a ReLU activation function in all experiments, while the decoder used a linear activation function, apart from the KDD example where the decoder's activation function was also ReLU. For the NSL KDD dataset, the encoder also had a batch normalization layer following the single dense layer. The attention layer of the MemAE models used in the experiments was set up with hard shrinkage of $\lambda = 1 / N$, where $N$ is the number of ``memory slots'' (rows) in the memory matrix. All experiments were run for a total of $25$ or $30$ global training rounds, having clients apply the Adam optimizer on their local copies of the model. A summary of these hyperparameters is listed out in Table~\ref{tab:hyperparams}.

\begin{table}[t]
    \caption{Experimental hyperparameters per dataset.}
    \label{tab:hyperparams}
    \centering
    \begin{tabular}{l | c c c}
    & \multicolumn{3}{c}{\textbf{Dataset}} \\
    \textbf{Parameter} & \textbf{KDD} & \textbf{NSL KDD} & \textbf{PAMAP2} \\
    \toprule
    Batch Size & 128 & 128 & 512 \\

    Encoding Dimension & 3 & 3 & 3 \\

    \# Global Training Steps & 30 & 30 & 25 \\
    
    Learning Rate & 0.001 & 0.01 & 0.001 \\
    
    \# Local Epochs & 10 & 10 & 10 \\
    
    Memory Dimension & 50 & 50 & 50 \\


    \bottomrule
    \end{tabular}
\end{table}

After each global training round, the FL server aggregates model updates from all clients through one of the above strategies, after which it validates the model on the test set. The validation is done by calculating the reconstruction error $e$ (Equation~\ref{eq:reconstruction_loss}) for all test data points, after which they are min-max normalized. An error threshold $\hat{e}$ is then chosen such that it maximises the $F1$ score of the model, as done in~\cite{10839838,11336743}. Points for which the reconstruction error is greater than $\hat{e}$ are deemed anomalies, while the rest are normal data points.

Training was conducted with full client participation, having $100$ clients in all experiments for each combination of dataset, model, and aggregation function.

\section{Results \& Discussion}
\label{sec:results}

An overview of the results is given in Table~\ref{tab:results}. Each model and aggregation strategy was trained and validated in both IID ($\alpha = 100$) and highly non-IID ($\alpha = 0.1$) scenarios. The table lists out the maximum recorded values for F1 score, ROC\_AUC, precision, and recall. 

The results were obtained by running the global model on a centralized dataset after each round of global training. Validation sets from all datasets had a $50:50$ split between outliers and inliers. We consider all of the following metrics, since their combination gives a comprehensive overview of the performance of each model under different circumstances.

All of the models performed well on the KDD dataset, with only slight variations. No matter the data skew, all aggregation strategies apart from the random one achieve F1 and ROC AUC scores above 99\%. The stochastic aggregation underperformed when data was more equally distributed ($\alpha = 100$). The regular autoencoder was the best model in the IID scenario, also coming in at a close second in the non-IID setup, just behind the K-Means approach to aggregating MemAE models. The KDD dataset has seen widespread use and modern techniques all achieve great results on it, which is why it is not viewed as much of a challenge as of late~\cite{10839838,DBLP:journals/corr/abs-2204-09825}. However, it is still an important baseline dataset ensuring validity and usability of newly developed techniques. 

The NSL KDD dataset was more challenging for all models, with all of them having an F1 score in the range of $93-94\%$, and ROC AUC between $97-98\%$. The facility location approach showed the best results in the more balanced setting, while for $\alpha = 0.1$, the best ROC AUC was recorded by the stochastic model at $98.08\%$. However, K-Means aggregation was not far behind in this regard, achieving a ROC AUC of $98.01\%$, while also recording the best F1 score of $94.39\%$. The regular autoencoder was outperformed by almost all of the MemAE models in terms of F1 and ROC AUC, for both balanced and unbalanced data distributions. It came in second-to-last only in terms of F1 score for $\alpha = 100$, outperforming the stochastic model by a slight margin.  

The PAMAP2 dataset proved to be the most challenging one.
In the balanced data distribution setting, the K-Means approach outperformed the MemAE aggregation using FedAvg in terms of F1 score by only $0.01\%$.
The highest ROC AUC of $88.28\%$ was achieved by the combination of FedAvg and MemAE. However, both the MemAE and stochastic models fell off when faced with extremely non-IID data. The facility location and K-Means approaches recorded F1 scores of $82.75\%$ and $81.64\%$, respectively. Both aggregation strategies were on top in terms of ROC AUC, achieving scores of $89.39\%$ and $88.42\%$.

\begin{table*}[t]
\caption{Experimental results for different $\alpha$ values.}
\label{tab:results}
\centering
\begin{tabular}{lll|cc|cc|cc|cc}
\toprule
 & & & \multicolumn{2}{c}{F1} & \multicolumn{2}{c}{ROC\_AUC} & \multicolumn{2}{c}{Precision} & \multicolumn{2}{c}{Recall} \\
 & & $\alpha$ & 0.1 & 100 & 0.1 & 100 & 0.1 & 100 & 0.1 & 100 \\

 Dataset & Aggregation & &  &  &  &  &  &  &  &  \\
\midrule
\multirow[t]{6}{*}{KDD} & autoencoder & & 99.21 & \textbf{99.30} & \textbf{99.66} & \textbf{99.75} & 99.87 & 99.91 & 98.57 & 98.68 \\
 & facility\_location & & 99.12 & 99.28 & 99.62 & 99.64 & \textbf{99.97} & \textbf{99.93} & 98.28 & 98.63 \\
 & kmeans & & \textbf{99.26} & 99.08 & 99.64 & 99.65 & 99.77 & 99.60 & \textbf{98.75} & 98.56 \\
 & kmedoids & & 99.04 & 99.25 & 99.44 & 99.62 & 99.47 & 99.87 & 98.62 & 98.64 \\
 & memae & & 99.25 & 99.25 & 99.64 & 99.65 & 99.86 & 99.79 & 98.66 & \textbf{98.72} \\
 & random & & 99.19 & 97.84 & 99.60 & 98.56 & 99.70 & 97.29 & 98.69 & 98.40 \\
\midrule
\multirow[t]{6}{*}{NSL KDD} & autoencoder & & 93.31 & 93.69 & 97.74 & 97.57 & 95.73 & 92.32 & 91.00 & \textbf{95.11} \\
 & facility\_location & & 93.74 & \textbf{94.39} & 97.96 & \textbf{98.58} & 94.36 & \textbf{97.14} & 93.13 & 91.79 \\
 & kmeans & & \textbf{94.39} & 94.27 & 98.01 & 98.44 & \textbf{96.64} & 94.58 & 92.25 & 93.96 \\
 & kmedoids & & 93.71 & 93.96 & 97.80 & 97.91 & 95.42 & 93.10 & 92.06 & 94.84 \\
 & memae & & 94.36 & 94.16 & 97.98 & 98.40 & 94.11 & 94.84 & \textbf{94.62} & 93.49 \\
 & random & & 94.23 & 93.53 & \textbf{98.08} & 98.22 & 94.77 & 94.03 & 93.70 & 93.04 \\
\midrule

\multirow[t]{6}{*}{PAMAP2} & autoencoder & & 80.94 & 77.99 & 86.95 & 86.46 & 74.37 & 67.46 & 88.78 & 92.41 \\
 & facility\_location & & \textbf{82.75} & 78.05 & \textbf{89.39} & 84.78 & 78.40 & 72.63 & 87.62 & 84.36 \\
 & kmeans & & 81.64 & \textbf{81.90} & 88.42 & 88.01 & 74.24 & 73.15 & \textbf{90.68} & 93.04 \\
 & kmedoids & & 79.40 & 78.04 & 87.25 & 84.78 & \textbf{79.15} & 73.45 & 79.66 & 83.25 \\
 & memae & & 77.66 & 81.89 & 83.69 & \textbf{88.28} & 78.41 & \textbf{74.39} & 76.92 & 91.07 \\
 & random & & 77.15 & 81.64 & 83.93 & 87.39 & 67.18 & 72.21 & 90.58 & \textbf{93.90} \\
\bottomrule
\end{tabular}
\end{table*}

Overall, there were no drastic differences in the KDD and NSL KDD experiments between the different models and aggregations, with results slightly favoring the novel aggregations. The guided memory aggregations, however, showed significant improvement over previous methods on the PAMAP2 dataset. These differences became apparent especially in the extremely non-IID case ($\alpha = 0.1$).

\begin{figure}[t]
    \centering
    \begin{subfigure}[t]{0.5\textwidth}
        \centering
        \includegraphics[width=\linewidth]{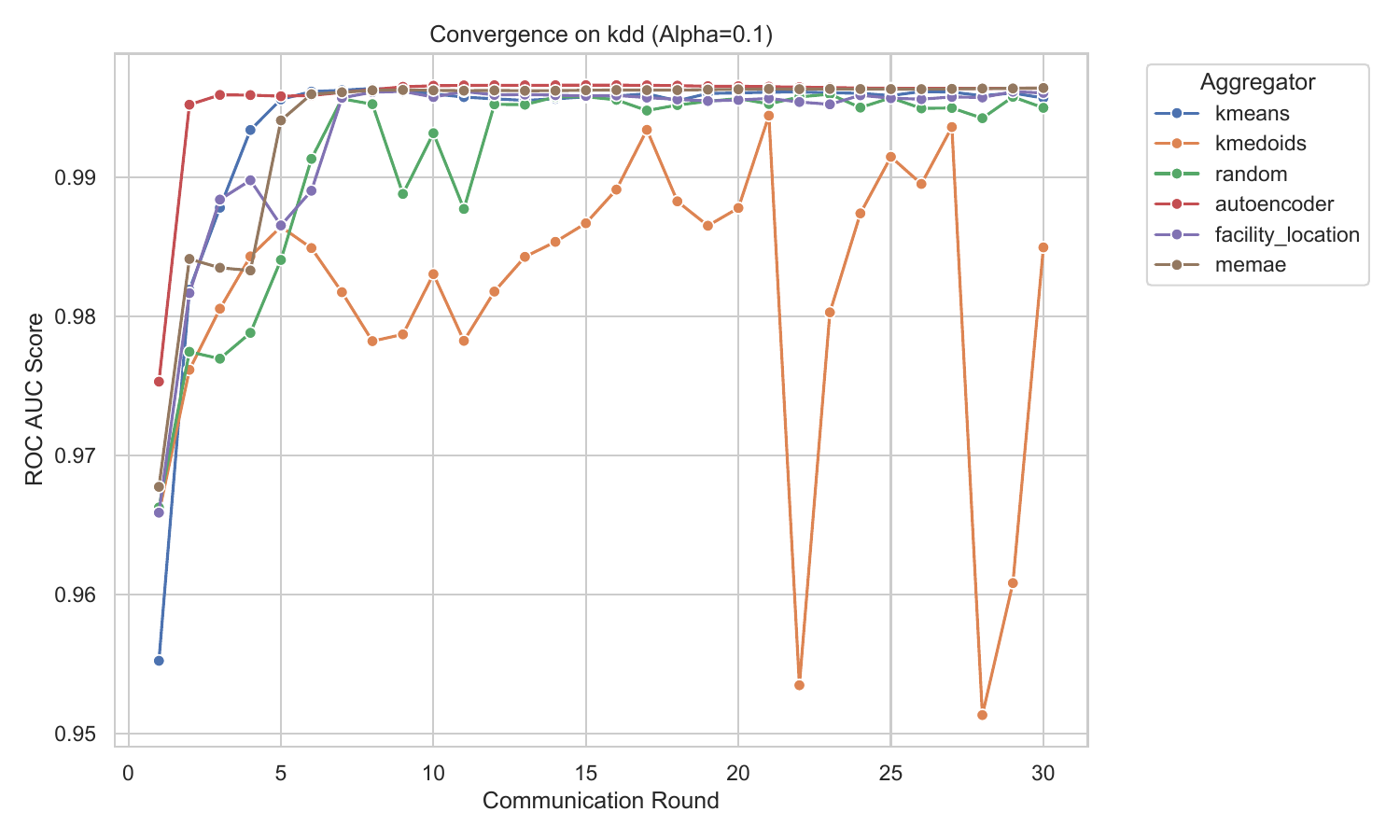}\\
        \caption{KDD ($\alpha = 0.1$)}
        \label{fig:learning-curves-kdd}
    \end{subfigure}\hfill
    \begin{subfigure}[t]{0.5\textwidth}
        \centering
        \includegraphics[width=\linewidth]{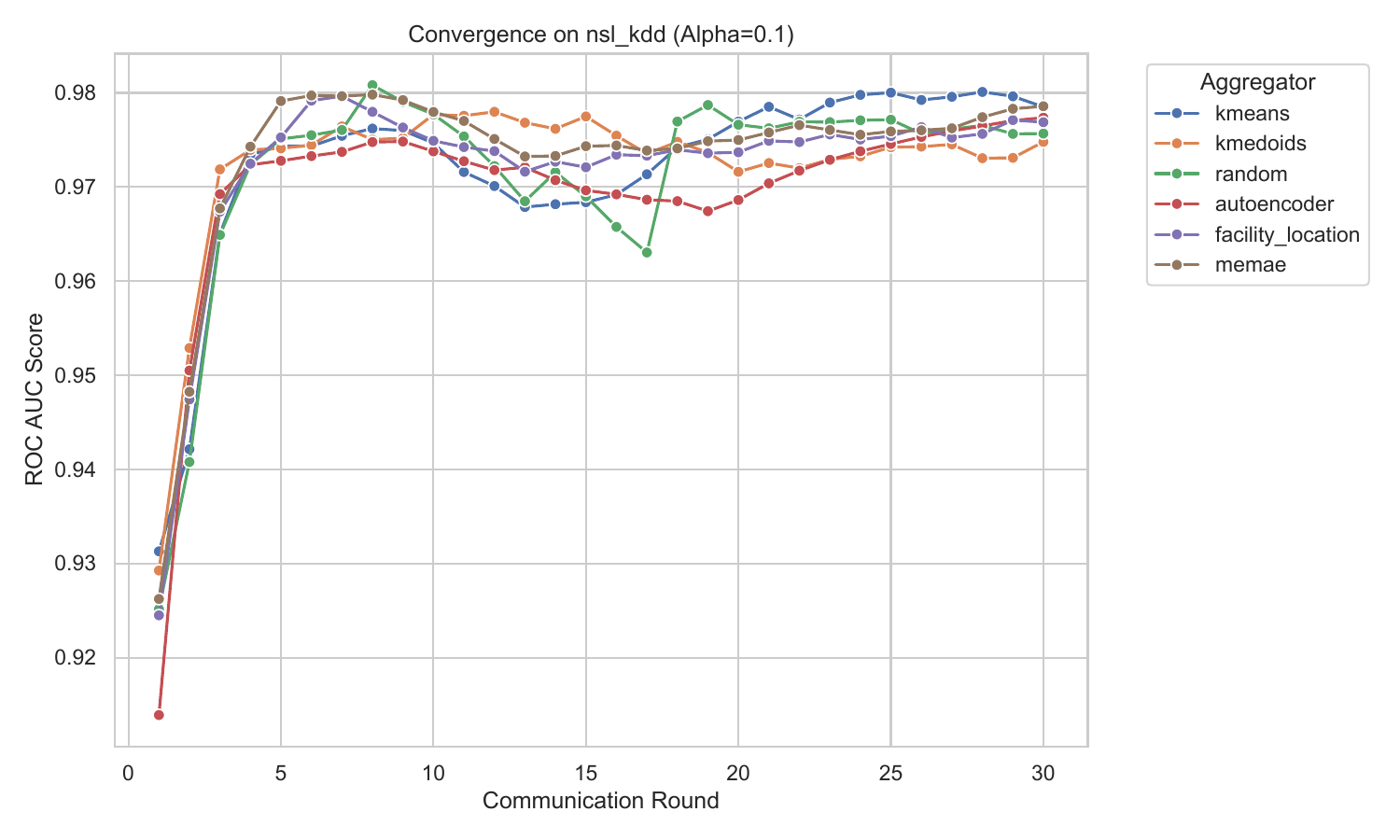}\\
        \caption{NSL KDD ($\alpha = 0.1$)}
        \label{fig:learning-curves-nsl-kdd}
    \end{subfigure}\hfill
    \begin{subfigure}[t]{0.5\textwidth}
        \centering
        \includegraphics[width=\linewidth]{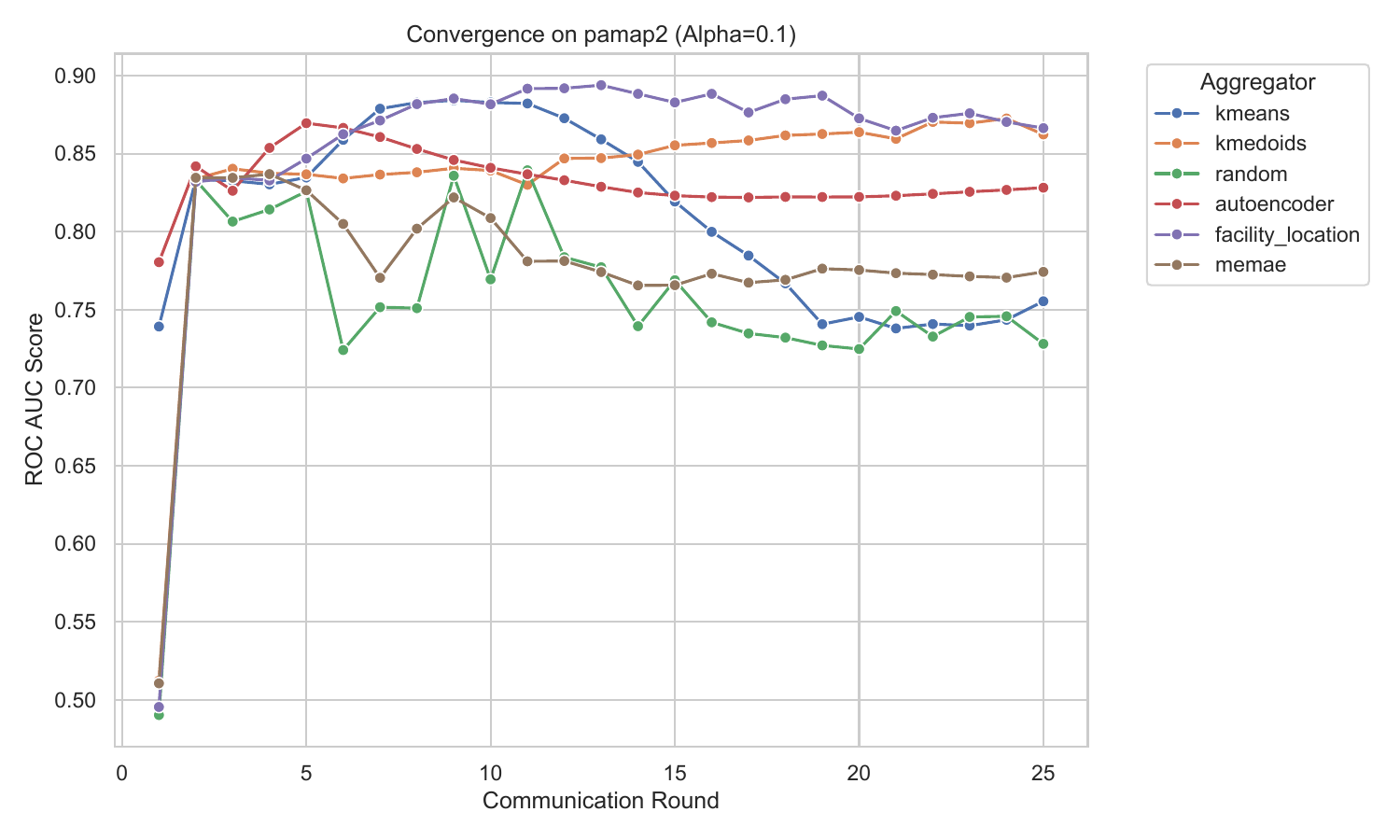}\\
        \caption{PAMAP2 ($\alpha = 0.1$)}
        \label{fig:learning-curves-pamap2}
    \end{subfigure}
    \caption{Learning curves based on ROC AUC for the three datasets in the highly non-IID setting ($\alpha = 0.1$).}
    \label{fig:learning-curves}
\end{figure}

Even though in some cases the scores of the different models were rather similar, it is worth noting how these models performed during training. Each round of global training was followed by centralized validation by the server. Results for the non-IID cases of the three datasets are pictured in Figure~\ref{fig:learning-curves}.

Figure~\ref{fig:learning-curves-kdd} depicts the validation scores while training on KDD. All of the models converged quite quickly, and remained stable. The K-Medoids and stochastic (random) aggregation strategies were the most unstable in this scenario. Training on NSL KDD was similar to the previous case, which can be seen in Figure~\ref{fig:learning-curves-nsl-kdd}. 
The random aggregation approach managed to record the best ROC AUC score in this particular scenario, however, when examining the validation curve during training, it is evident that it is the most volatile model. This volatility allowed it to secure the best result, but is also a concern and a sign of unreliability in the approach. 

Selection-based aggregations (K-Medoids and Facility Location) had the steadiest performance in the PAMAP2 case (Figure~\ref{fig:learning-curves-pamap2}). Random aggregation struggled to find its way, being quite volatile in performance during learning. Using simple FedAvg on the MemAE model also did not yield substantially good results, and the regular autoencoder did not manage to capture representations good enough to distinguish between inliers and outliers. The K-Means approach started off well, but fell off significantly as training went along. 

Based on these results, the guided approaches to aggregating memory-augmented autoencoders do provide certain advantages over simply aggregating the memory layer using standard approaches, without taking into account the intricacies of the underlying architecture of this model. The attention based models remain effective in federated settings, achieving comparable or even improved performance over regular autoencoders. Finally, stochastic aggregation is a valid approach to combining memory modules, however, due to its volatility, it is an unreliable method compared to the guided approaches validated in these experiments.

\section{Conclusion}

Attention layers are a powerful mechanism driving machine learning innovation in numerous fields. One such domain is federated learning. In this paper we build on previous research~\cite{Gong_2019_ICCV, 10839838, 11336743}, leveraging representation learning and attention mechanisms to obtain robust models for federated outlier detection. The MemAE architecture coupled with the newly-proposed methods in this paper allow training of robust anomaly detection models in federated environments.

We propose novel, guided, federated aggregation functions tailored specifically for memory layers, such as those found in the MemAE architecture~\cite{Gong_2019_ICCV}. We validate our approach on three different datasets in six different scenarios, and compare the novel functions to regular autoencoders, MemAE aggregation using FedAvg~\cite{10839838}, and previously proposed stochastic row-wise aggregations of the attention layers~\cite{11336743}. Our novel methods achieve comparable results on established outlier detection datasets, and outperform other baseline methods in challenging non-IID scenarios. We take into account the intrinsics of the MemAE architecture, and achieve convergence more steadily than the previous stochastic approach. We validated our approach using a single layer encoder and decoder architecture, ensuring this method can be applied using minimal computational requirements, meaning that it is suitable for massively distributed IoT environments.

Federated anomaly detection would benefit greatly from standardized benchmark datasets like FedAD-Bench~\cite{10839838}, specifically designed for these types of problems. We propose building on these benchmark initiatives by extending the collection of datasets to include a broader selection of data modalities, and by balancing the ratio of inliers and outliers in the validation set such as in the experiments presented in this paper.

In future work, we aim to conduct a more in-depth examination of the properties of the distributed dataset, and their effect on the performance of the different guided aggregation approaches.
Specifically, we plan to answer the question of which circumstances determine whether it is better to rely on a selection-based approach like facility location or one based on aggregated results like K-Means, where new memory entries are constructed (centroids) based on clusters of submitted memory candidates.

\section*{Acknowledgment}

This paper has been supported by the European Union’s Horizon Europe research and innovation actions under grant agreements No 101135775 (PANDORA) and No 101215032 (TURING). 

The authors from the University of Novi Sad gratefully acknowledge the financial support of the Ministry of Science, Technological Development and Innovation of the Republic of Serbia (Grants No. 451-03-137/2025-03/ 200125 \& 451-03-136/2025-03/ 200125).

\bibliographystyle{abbrv}
\bibliography{ref}

@article{10.1561/2200000083,
    author = {Kairouz, Peter and McMahan, H. Brendan},
    title = {Advances and Open Problems in Federated Learning},
    journal = {Foundations and Trends in Machine Learning},
    volume = {14},
    number = {1-2},
    pages = {1-210},
    year = {2021},
    month = {06},
    issn = {1935-8237},
    doi = {10.1561/2200000083},
    url = {https://doi.org/10.1561/2200000083},
    eprint = {https://www.emerald.com/ftmal/article-pdf/14/1-2/1/11147179/2200000083en.pdf}
}

@InProceedings{pmlr-v54-mcmahan17a,
  title = 	 {{Communication-Efficient Learning of Deep Networks from Decentralized Data}},
  author = 	 {McMahan, Brendan and Moore, Eider and Ramage, Daniel and Hampson, Seth and Arcas, Blaise Aguera y},
  booktitle = 	 {Proceedings of the 20th International Conference on Artificial Intelligence and Statistics},
  pages = 	 {1273--1282},
  year = 	 {2017},
  editor = 	 {Singh, Aarti and Zhu, Jerry},
  volume = 	 {54},
  series = 	 {Proceedings of Machine Learning Research},
  month = 	 {20--22 Apr},
  publisher =    {PMLR},
  url = 	 {https://proceedings.mlr.press/v54/mcmahan17a.html}
}

@article{ilic2024towards,
  title={Towards optimal learning: Investigating the impact of different model updating strategies in federated learning},
  author={Ili{\'c}, Mihailo and Ivanovi{\'c}, Mirjana and Kurbalija, Vladimir and Valachis, Antonios},
  journal={Expert Systems with Applications},
  volume={249},
  pages={123553},
  year={2024},
  publisher={Elsevier}
}

@InProceedings{10.1007/978-3-030-11723-8_9,
author="Sheller, Micah J.
and Reina, G. Anthony
and Edwards, Brandon
and Martin, Jason
and Bakas, Spyridon",
editor="Crimi, Alessandro
and Bakas, Spyridon
and Kuijf, Hugo
and Keyvan, Farahani
and Reyes, Mauricio
and van Walsum, Theo",
title="Multi-institutional Deep Learning Modeling Without Sharing Patient Data: A Feasibility Study on Brain Tumor Segmentation",
booktitle="Brainlesion: Glioma, Multiple Sclerosis, Stroke and Traumatic Brain Injuries",
year="2019",
publisher="Springer International Publishing",
address="Cham",
pages="92--104",
isbn="978-3-030-11723-8"
}

@inproceedings{MLSYS2020_1f5fe839,
 author = {Li, Tian and Sahu, Anit Kumar and Zaheer, Manzil and Sanjabi, Maziar and Talwalkar, Ameet and Smith, Virginia},
 booktitle = {Proceedings of Machine Learning and Systems},
 editor = {I. Dhillon and D. Papailiopoulos and V. Sze},
 pages = {429--450},
 title = {Federated Optimization in Heterogeneous Networks},
 url = {https://proceedings.mlsys.org/paper\_files/paper/2020/file/1f5fe83998a09396ebe6477d9475ba0c-Paper.pdf},
 volume = {2},
 year = {2020}
}

@article{niu2023fedhybrid,
  title={FedHybrid: A hybrid federated optimization method for heterogeneous clients},
  author={Niu, Xiaochun and Wei, Ermin},
  journal={IEEE Transactions on Signal Processing},
  volume={71},
  pages={150--163},
  year={2023},
  publisher={IEEE}
}

@INPROCEEDINGS{11336743,
  author={Ilić, Mihailo and Savić, Miloš and Kurbalija, Vladimir and Ivanović, Mirjana and Fortino, Giancarlo and Jakovetić, Dušan},
  booktitle={2025 3rd International Conference on Federated Learning Technologies and Applications (FLTA)}, 
  title={Federated Attention Autoencoders with a Stochastic Aggregation Scheme for Anomaly Detection}, 
  year={2025},
  volume={},
  number={},
  pages={270-277},
  doi={10.1109/FLTA67013.2025.11336743}
}

@InProceedings{Gong_2019_ICCV,
author = {Gong, Dong and Liu, Lingqiao and Le, Vuong and Saha, Budhaditya and Mansour, Moussa Reda and Venkatesh, Svetha and Hengel, Anton van den},
title = {Memorizing Normality to Detect Anomaly: Memory-Augmented Deep Autoencoder for Unsupervised Anomaly Detection},
booktitle = {Proceedings of the IEEE/CVF International Conference on Computer Vision (ICCV)},
month = {October},
year = {2019}
}

@INPROCEEDINGS{10839838,
  author={Anwar, Ahmed and Moser, Brian and Herurkar, Dayananda and Raue, Federico and Hegiste, Vinit and Legler, Tatjana and Dengel, Andreas},
  booktitle={2024 2nd International Conference on Federated Learning Technologies and Applications (FLTA)}, 
  title={FedAD-Bench: A Unified Benchmark for Federated Unsupervised Anomaly Detection in Tabular Data}, 
  year={2024},
  volume={},
  number={},
  pages={115-122},
  doi={10.1109/FLTA63145.2024.10839838}
}

@misc{caldas2019leafbenchmarkfederatedsettings,
      title={LEAF: A Benchmark for Federated Settings}, 
      author={Sebastian Caldas and Sai Meher Karthik Duddu and Peter Wu and Tian Li and Jakub Konečný and H. Brendan McMahan and Virginia Smith and Ameet Talwalkar},
      year={2019},
      eprint={1812.01097},
      archivePrefix={arXiv},
      primaryClass={cs.LG},
      url={https://arxiv.org/abs/1812.01097}
}

@misc{kdd_cup_1999_data_130,
  author       = {Stolfo, Salvatore and Fan, Wei and Lee, Wenke and Prodromidis, Andreas and Chan, Philip},
  title        = {{KDD Cup 1999 Data}},
  year         = {1999},
  howpublished = {UCI Machine Learning Repository},
  note         = {{DOI}: https://doi.org/10.24432/C51C7N}
}

@inproceedings{tavallaee2009detailed,
  title={A detailed analysis of the KDD CUP 99 data set},
  author={Tavallaee, Mahbod and Bagheri, Ebrahim and Lu, Wei and Ghorbani, Ali A},
  booktitle={2009 IEEE symposium on computational intelligence for security and defense applications},
  pages={1--6},
  year={2009},
  organization={Ieee}
}

@misc{pamap2_physical_activity_monitoring_231,
  author       = {Reiss, Attila},
  title        = {{PAMAP2 Physical Activity Monitoring}},
  year         = {2012},
  howpublished = {UCI Machine Learning Repository},
  note         = {{DOI}: https://doi.org/10.24432/C5NW2H}
}

@misc{hsu2019measuringeffectsnonidenticaldata,
      title={Measuring the Effects of Non-Identical Data Distribution for Federated Visual Classification}, 
      author={Tzu-Ming Harry Hsu and Hang Qi and Matthew Brown},
      year={2019},
      eprint={1909.06335},
      archivePrefix={arXiv},
      primaryClass={cs.LG},
      url={https://arxiv.org/abs/1909.06335}, 
}

@article{krause2014submodular,
  title={Submodular function maximization.},
  author={Krause, Andreas and Golovin, Daniel},
  journal={Tractability},
  volume={3},
  number={71-104},
  pages={3},
  year={2014}
}

@article{JMLR:v21:19-467,
  author  = {Jacob Schreiber and Jeffrey Bilmes and William Stafford Noble},
  title   = {apricot: Submodular selection for data summarization in Python},
  journal = {Journal of Machine Learning Research},
  year    = {2020},
  volume  = {21},
  number  = {161},
  pages   = {1--6},
  url     = {http://jmlr.org/papers/v21/19-467.html}
}

@article{DBLP:journals/access/NassifAND21,
  author       = {Ali Bou Nassif and
                  Manar Abu Talib and
                  Qassim Nasir and
                  Fatima Mohamad Dakalbab},
  title        = {Machine Learning for Anomaly Detection: {A} Systematic Review},
  journal      = {{IEEE} Access},
  volume       = {9},
  pages        = {78658--78700},
  year         = {2021},
  url          = {https://doi.org/10.1109/ACCESS.2021.3083060},
  doi          = {10.1109/ACCESS.2021.3083060},
  bibsource    = {dblp computer science bibliography, https://dblp.org}
}

@article{DBLP:journals/access/XiaCYM23,
  author       = {Geming Xia and
                  Jian Chen and
                  Chaodong Yu and
                  Jun Ma},
  title        = {Poisoning Attacks in Federated Learning: {A} Survey},
  journal      = {{IEEE} Access},
  volume       = {11},
  pages        = {10708--10722},
  year         = {2023},
  url          = {https://doi.org/10.1109/ACCESS.2023.3238823},
  doi          = {10.1109/ACCESS.2023.3238823},
  bibsource    = {dblp computer science bibliography, https://dblp.org}
}

@article{DBLP:journals/air/ZhangYMN24,
  author       = {Chang Zhang and
                  Shunkun Yang and
                  Lingfeng Mao and
                  Huansheng Ning},
  title        = {Anomaly detection and defense techniques in federated learning: a
                  comprehensive review},
  journal      = {Artif. Intell. Rev.},
  volume       = {57},
  number       = {6},
  pages        = {150},
  year         = {2024},
  url          = {https://doi.org/10.1007/s10462-024-10796-1},
  doi          = {10.1007/S10462-024-10796-1},
  bibsource    = {dblp computer science bibliography, https://dblp.org}
}

@article{DBLP:journals/corr/abs-2204-09825,
  author       = {Maxime Alvarez and
                  Jean{-}Charles Verdier and
                  D'Jeff K. Nkashama and
                  Marc Frappier and
                  Pierre{-}Martin Tardif and
                  Froduald Kabanza},
  title        = {A Revealing Large-Scale Evaluation of Unsupervised Anomaly Detection
                  Algorithms},
  journal      = {CoRR},
  volume       = {abs/2204.09825},
  year         = {2022},
  url          = {https://doi.org/10.48550/arXiv.2204.09825},
  doi          = {10.48550/ARXIV.2204.09825},
  eprinttype   = {arXiv},
  eprint       = {2204.09825},
  bibsource    = {dblp computer science bibliography, https://dblp.org}
}

@inproceedings{DBLP:conf/iclr/Balakrishnan0ZH22,
  author       = {Ravikumar Balakrishnan and
                  Tian Li and
                  Tianyi Zhou and
                  Nageen Himayat and
                  Virginia Smith and
                  Jeff A. Bilmes},
  title        = {Diverse Client Selection for Federated Learning via Submodular Maximization},
  booktitle    = {The Tenth International Conference on Learning Representations, {ICLR}
                  2022, Virtual Event, April 25-29, 2022},
  publisher    = {OpenReview.net},
  year         = {2022},
  url          = {https://openreview.net/forum?id=nwKXyFvaUm},
  bibsource    = {dblp computer science bibliography, https://dblp.org}
}

@inproceedings{DBLP:conf/isc2/ArmackiBJK22,
  author       = {Aleksandar Armacki and
                  Dragana Bajovic and
                  Dusan Jakovetic and
                  Soummya Kar},
  title        = {Personalized Federated Learning via Convex Clustering},
  booktitle    = {{IEEE} International Smart Cities Conference, {ISC2} 2022, Pafos,
                  Cyprus, September 26-29, 2022},
  pages        = {1--7},
  publisher    = {{IEEE}},
  year         = {2022},
  url          = {https://doi.org/10.1109/ISC255366.2022.9921863},
  doi          = {10.1109/ISC255366.2022.9921863},
  bibsource    = {dblp computer science bibliography, https://dblp.org}
}

@article{DBLP:journals/bigdatama/XiongZSWWLG24,
  author       = {Ao Xiong and
                  Han Zhou and
                  Yu Song and
                  Dong Wang and
                  Xu Wei and
                  Da Li and
                  Bo Gao},
  title        = {A Multi-Task Based Clustering Personalized Federated Learning Method},
  journal      = {Big Data Min. Anal.},
  volume       = {7},
  number       = {4},
  pages        = {1017--1030},
  year         = {2024},
  url          = {https://doi.org/10.26599/bdma.2024.9020001},
  doi          = {10.26599/BDMA.2024.9020001},
  bibsource    = {dblp computer science bibliography, https://dblp.org}
}

@INPROCEEDINGS{9644782,
  author={Duan, Moming and Liu, Duo and Ji, Xinyuan and Liu, Renping and Liang, Liang and Chen, Xianzhang and Tan, Yujuan},
  booktitle={2021 IEEE Intl Conf on Parallel \& Distributed Processing with Applications, Big Data \& Cloud Computing, Sustainable Computing \& Communications, Social Computing \& Networking (ISPA/BDCloud/SocialCom/SustainCom)}, 
  title={FedGroup: Efficient Federated Learning via Decomposed Similarity-Based Clustering}, 
  year={2021},
  volume={},
  number={},
  pages={228-237},
  doi={10.1109/ISPA-BDCloud-SocialCom-SustainCom52081.2021.00042}}

@ARTICLE{9743558,
  author={Tan, Alysa Ziying and Yu, Han and Cui, Lizhen and Yang, Qiang},
  journal={IEEE Transactions on Neural Networks and Learning Systems}, 
  title={Towards Personalized Federated Learning}, 
  year={2023},
  volume={34},
  number={12},
  pages={9587-9603},
  doi={10.1109/TNNLS.2022.3160699}}

@article{10.1145/3638052,
author = {Zhang, Jinghui and Wang, Jiawei and Li, Yaning and Xin, Fa and Dong, Fang and Luo, Junzhou and Wu, Zhihua},
title = {Addressing Heterogeneity in Federated Learning with Client Selection via Submodular Optimization},
year = {2024},
issue_date = {March 2024},
publisher = {Association for Computing Machinery},
address = {New York, NY, USA},
volume = {20},
number = {2},
issn = {1550-4859},
url = {https://doi.org/10.1145/3638052},
doi = {10.1145/3638052},
journal = {ACM Trans. Sen. Netw.},
month = feb,
articleno = {48},
numpages = {32}
}

@inproceedings{NIPS2013_a1d50185,
 author = {Iyer, Rishabh K and Bilmes, Jeff A},
 booktitle = {Advances in Neural Information Processing Systems},
 editor = {C.J. Burges and L. Bottou and M. Welling and Z. Ghahramani and K. Weinberger},
 pages = {},
 publisher = {Curran Associates, Inc.},
 title = {Submodular Optimization with Submodular Cover and Submodular Knapsack Constraints},
 url = {https://proceedings.neurips.cc/paper_files/paper/2013/file/a1d50185e7426cbb0acad1e6ca74b9aa-Paper.pdf},
 volume = {26},
 year = {2013}
}

@article{DBLP:journals/www/LongXSZWJ23,
  author       = {Guodong Long and
                  Ming Xie and
                  Tao Shen and
                  Tianyi Zhou and
                  Xianzhi Wang and
                  Jing Jiang},
  title        = {Multi-center federated learning: clients clustering for better personalization},
  journal      = {World Wide Web {(WWW)}},
  volume       = {26},
  number       = {1},
  pages        = {481--500},
  year         = {2023},
  url          = {https://doi.org/10.1007/s11280-022-01046-x},
  doi          = {10.1007/S11280-022-01046-X},
  bibsource    = {dblp computer science bibliography, https://dblp.org}
}

@article{DBLP:journals/tnn/LyuYMCSZYY24,
  author       = {Lingjuan Lyu and
                  Han Yu and
                  Xingjun Ma and
                  Chen Chen and
                  Lichao Sun and
                  Jun Zhao and
                  Qiang Yang and
                  Philip S. Yu},
  title        = {Privacy and Robustness in Federated Learning: Attacks and Defenses},
  journal      = {{IEEE} Trans. Neural Networks Learn. Syst.},
  volume       = {35},
  number       = {7},
  pages        = {8726--8746},
  year         = {2024},
  url          = {https://doi.org/10.1109/TNNLS.2022.3216981},
  doi          = {10.1109/TNNLS.2022.3216981},
  bibsource    = {dblp computer science bibliography, https://dblp.org}
}

@article{DBLP:journals/access/NetoHDMF23,
  author       = {H{\'{e}}lio N. Cunha Neto and
                  Jernej Hribar and
                  Ivana Dusparic and
                  Diogo Menezes Ferrazani Mattos and
                  Natalia Castro Fernandes},
  title        = {A Survey on Securing Federated Learning: Analysis of Applications,
                  Attacks, Challenges, and Trends},
  journal      = {{IEEE} Access},
  volume       = {11},
  pages        = {41928--41953},
  year         = {2023},
  url          = {https://doi.org/10.1109/ACCESS.2023.3269980},
  doi          = {10.1109/ACCESS.2023.3269980},
  bibsource    = {dblp computer science bibliography, https://dblp.org}
}

@article{DBLP:journals/access/JithishAMY23,
  author       = {J. Jithish and
                  Bithin Alangot and
                  Nagarajan Mahalingam and
                  Kiat Seng Yeo},
  title        = {Distributed Anomaly Detection in Smart Grids: {A} Federated Learning-Based
                  Approach},
  journal      = {{IEEE} Access},
  volume       = {11},
  pages        = {7157--7179},
  year         = {2023},
  url          = {https://doi.org/10.1109/ACCESS.2023.3237554},
  doi          = {10.1109/ACCESS.2023.3237554},
  bibsource    = {dblp computer science bibliography, https://dblp.org}
}

@INPROCEEDINGS{11336407,
  author={Herurkar, Dayananda and Anwar, Ahmed and Palacio, Sebastian and Hees, Jörn and Dengel, Andreas},
  booktitle={2025 3rd International Conference on Federated Learning Technologies and Applications (FLTA)}, 
  title={Fin-Fed-OD: Enhancing Outlier Detection Using Federated Learning on Financial Tabular Data}, 
  year={2025},
  volume={},
  number={},
  pages={40-47},
  doi={10.1109/FLTA67013.2025.11336407}}

@ARTICLE{10566052,
  author={Kong, Xiangjie and Zhang, Wenyi and Wang, Hui and Hou, Mingliang and Chen, Xin and Yan, Xiaoran and Das, Sajal K.},
  journal={IEEE Transactions on Neural Networks and Learning Systems}, 
  title={Federated Graph Anomaly Detection via Contrastive Self-Supervised Learning}, 
  year={2025},
  volume={36},
  number={5},
  pages={7931-7944},
  doi={10.1109/TNNLS.2024.3414326}}

@ARTICLE{6472238,
  author={Bengio, Yoshua and Courville, Aaron and Vincent, Pascal},
  journal={IEEE Transactions on Pattern Analysis and Machine Intelligence}, 
  title={Representation Learning: A Review and New Perspectives}, 
  year={2013},
  volume={35},
  number={8},
  pages={1798-1828},
  doi={10.1109/TPAMI.2013.50}}

@inproceedings{10.1145/3378679.3394530,
author = {van Berlo, Bram and Saeed, Aaqib and Ozcelebi, Tanir},
title = {Towards federated unsupervised representation learning},
year = {2020},
isbn = {9781450371322},
publisher = {Association for Computing Machinery},
address = {New York, NY, USA},
url = {https://doi.org/10.1145/3378679.3394530},
doi = {10.1145/3378679.3394530},
booktitle = {Proceedings of the Third ACM International Workshop on Edge Systems, Analytics and Networking},
pages = {31–36},
numpages = {6},
location = {Heraklion, Greece},
series = {EdgeSys '20}
}

@ARTICLE{9492755,
  author={Mills, Jed and Hu, Jia and Min, Geyong},
  journal={IEEE Transactions on Parallel and Distributed Systems}, 
  title={Multi-Task Federated Learning for Personalised Deep Neural Networks in Edge Computing}, 
  year={2022},
  volume={33},
  number={3},
  pages={630-641},
  doi={10.1109/TPDS.2021.3098467}}

@article{10.1145/3439950,
author = {Pang, Guansong and Shen, Chunhua and Cao, Longbing and Hengel, Anton Van Den},
title = {Deep Learning for Anomaly Detection: A Review},
year = {2021},
issue_date = {March 2022},
publisher = {Association for Computing Machinery},
address = {New York, NY, USA},
volume = {54},
number = {2},
issn = {0360-0300},
url = {https://doi.org/10.1145/3439950},
doi = {10.1145/3439950},
journal = {ACM Comput. Surv.},
month = mar,
articleno = {38},
numpages = {38}
}

@article{SAVIC2022116409,
title = {{Tax evasion risk management using a Hybrid Unsupervised Outlier Detection method}},
journal = {Expert Systems with Applications},
volume = {193},
pages = {116409},
year = {2022},
issn = {0957-4174},
doi = {https://doi.org/10.1016/j.eswa.2021.116409},
url = {https://www.sciencedirect.com/science/article/pii/S0957417421016973},
author = {Miloš Savić and Jasna Atanasijević and Dušan Jakovetić and Nataša Krejić}
}

@ARTICLE{9402912,
  author={Savic, Milos and Lukic, Milan and Danilovic, Dragan and Bodroski, Zarko and Bajović, Dragana and Mezei, Ivan and Vukobratovic, Dejan and Skrbic, Srdjan and Jakovetić, Dusan},
  journal={IEEE Access}, 
  title={{Deep Learning Anomaly Detection for Cellular IoT With Applications in Smart Logistics}}, 
  year={2021},
  volume={9},
  number={},
  pages={59406-59419},
  doi={10.1109/ACCESS.2021.3072916}
}

@article{10.1145/3381028,
author = {Boukerche, Azzedine and Zheng, Lining and Alfandi, Omar},
title = {Outlier Detection: Methods, Models, and Classification},
year = {2020},
issue_date = {May 2021},
publisher = {Association for Computing Machinery},
address = {New York, NY, USA},
volume = {53},
number = {3},
issn = {0360-0300},
url = {https://doi.org/10.1145/3381028},
doi = {10.1145/3381028},
journal = {ACM Comput. Surv.},
month = jun,
articleno = {55},
numpages = {37}
}

@ARTICLE{RufPIEEE21,
  author = {Lukas Ruff and Jacob R. Kauffmann and Robert A. Vandermeulen and Gr{\'e}goire Montavon and Wojciech Samek and Marius Kloft and Thomas G. Dietterich and Klaus-Robert M{\"u}ller},
  title = {A Unifying Review of Deep and Shallow Anomaly Detection},
  journal = {Proceedings of the IEEE},
  year = {2021},
  volume = {109},
  number = {5},
  pages = {756-795},
  doi = {10.1109/JPROC.2021.3052449},
  url = {https://doi.org/10.1109/JPROC.2021.3052449}
}

\end{document}